# Automatic Mammogram image Breast Region Extraction and Removal of Pectoral Muscle

R. Subash Chandra Boss, K. Thangavel, D. Arul Pon Daniel

**Abstract**— Currently Mammography is a most effective imaging modality used by radiologists for the screening of breast cancer. Finding an accurate, robust and efficient breast region segmentation technique still remains a challenging problem in digital mammography. Extraction of the breast profile region and the removal of pectoral muscle are essential pre-processing steps in Computer Aided Diagnosis (CAD) system for the diagnosis of breast cancer. Primarily it allows the search for abnormalities to be limited to the region of the breast tissue without undue influence from the background of the mammogram. The presence of pectoral muscle in mammograms biases detection procedures, which recommends removing the pectoral muscle during mammogram image pre-processing. The presence of pectoral muscle in mammograms may disturb or influence the detection of breast cancer as the pectoral muscle and mammographic parenchymas appear similar. The goal of breast region extraction is reducing the image size without losing anatomic information, it improve the accuracy of the overall CAD system. The main objective of this study is to propose an automated method to identify the pectoral muscle in Medio-Lateral Oblique (MLO) view mammograms. In this paper, we proposed histogram based 8-neighborhood connected component labelling method for breast region extraction and removal of pectoral muscle. The proposed method is evaluated by using the mean values of accuracy and error. The comparative analysis shows that the proposed method identifies the breast region more accurately.

—————— ◆ ——————

## 1 INTRODUCTION

Breast cancer has emerged as the leading sites of cancer in India among males and females. It is the most common type and frequent form of cancer and one in 22 women in India is likely to suffer from breast cancer. This is the second main cause of cancer deaths among women. Recently, the estimated number of Breast cancer cases for the years 2015 and 2020 will be 106,124 and 123,634 respectively, according to the National Cancer Registry Programme report of the Indian Council of Medical Research (ICMR). Detecting a cancer at an early stage can improve the cure rate from breast cancer [1].

Digital Mammogram is one of the important methods to identify the Breast Cancer at an early stage at some extend. The advantages of digital mammography include the lack of ionizing radiation, its non-invasiveness, the relatively compact instrumentation, and its cost-effectiveness. While mammography has been proven to be the most effective and reliable method for the early detection of breast cancer, as indicated by Siddiqui et al.[2], the large number of mammograms, generated by population screening, must be interpreted and diagnosed by a relatively small number of radiologists. In addition, when observing a mammographic image, abnormalities are often embedded in and camouflaged by varying densities of breast tissue structures, resulting in high rates of missed breast cancer cases as mentioned by Wroblewska et al. [3].

In order to reduce the increasing work-load and improve the accuracy of interpreting mammograms, a variety of CAD systems, that perform computerized mammographic analysis have been proposed, as stated by Rangayyan et al.[4].

These systems are usually employed as a second reader, with the final decision regarding the presence of a cancer left to the radiologist. Thus, their role in modern medical practice is considered to be significant and important in the early detection of breast cancer.

All of the CAD systems require, as a first stage, the segmentation of each mammogram into its representative anatomical regions, i.e., the breast border and pectoral muscle. The breast border extraction is a necessary and cumbersome step for typical CAD systems, as it must identify the breast region independently of the digitization system, the orientation of the breast in the image and the presence of noise, including imaging artifacts. The goal is to exclude the background from the subsequent processing steps, reducing the image file size without losing anatomic information. It should also have a fast running time and be sufficiently precise, in order to improve the accuracy of the overall CAD system [20, 21].

This paper is organized as follows: Section 2 discusses the related work. Section 3 discusses the Pre-processing. Section 4 discusses the proposed approach. Section 5 describes the worked example. Section 6 discusses experimental results and Section 7 covers conclusion.

## 2 RELATED WORKS

One of the most used pectoral muscle segmentation algorithms is the method proposed by Ferrari et al. [5] based on the Hough transform. The main problem with this approach is that the pectoral muscle is approximated by a line. These

————————————
- R. Subash Chandra Boss, is currently pursuing Ph.D program in Computer Science in Periyar University,Salem, Tamilnadu, India, PH-+91 90258 09892. E-mail: rmsubash_18@yahoo.co.in
- Dr. K. Thangavel is currently working as HOD, Department of Computer Science in Periyar University,Salem, Tamilnadu, India, PH-+91 94863 37009. E-mail: ktvelu@yahoo.com
- D. Arul Pon Daniel, is currently pursuing Ph.D program in Computer Science, Periyar University,Salem, Tamilnadu, India, PH-+9194437 49695. E-mail: apdaniel86@yahoo.com

methods give poor results when the pectoral muscle contour is a curve. For this reason, the same team proposed another method [6] based on Gabor wavelets. In [7], the pectoral muscle was once again approximated by a straight line, but this line was further adjusted through surface smoothing and edge detection.Ma et al. [8] described two image segmentation methods: one based on adaptive pyramids and other based on minimum spanning trees. The article [9] chose the longest straight line in Radon-domain as an approximation to the pectoral muscle localization. The problem with this work is twofold: the simplification of using a line and the use of a private database, so the results cannot be compared with other publications. Camilus and co-workers [10] used a graph cut method followed by Bezier curve smoothing. Recently, an isocontour map methodology was proposed [11]. Finally, a discrete time Markov chain and an active contour model were adopted in [12] for muscle detection.

Although the long list of related works, none addresses the problem of deciding if the muscle contour is present or not in the mammogram, all assuming that it is. Moreover, some works assume user input, either in the form of a region of interest or as a set of points in the contour. Finally, with the increased use of digital mammograms, and with its inherent higher quality, simpler approaches could be more adequate.

## 3 PRE-PROCESSING

### 3.1 Normalization

Normalization is a process that changes the range of pixel intensity values. For instance, applications include photographs with poor contrast due to glare. Normalization is sometimes called contrast stretching. In more general fields of data processing, such as digital signal processing, it is referred to as dynamic range expansion. The purpose of dynamic range expansion in the various applications is usually to bring the image, or other type of signal, into a range that is more familiar or normal to the senses, hence the term normalization [15, 23]. Normalization is a linear process. If the intensity range of the image is 30 to 230 and the desired range is 0 to 255 the process entails subtracting 50 from each of pixel intensity, making the range 0 to 200. Then each pixel intensity is multiplied by 255/200, making the range 0 to 255. Auto-normalization in image processing software typically normalizes to the full dynamic range of the number system specified in the image file format. The normalization process will produce iris regions, which have the same constant dimensions, so that two photographs of the same iris under different conditions will have characteristic features at the same spatial location.

### 3.2 Gray-level Normalization of Image

The distribution of gray levels of breast mammogram images may vary greatly; however, the ranges of the intensities are narrow. Normalization is a necessary step, and we normalize the mammogram image by mapping the intensity levels into the range [gmin, gmax]. The gray level normalization formula given below.

$$g(i,j) = g_{min} + \frac{(g_{max} - g_{min}) \mathrm{X} (g_o(i,j) - g_{o\,min})}{(g_{o\,max} - g_{o\,min})}$$

where go min and go max are the minimum and maximum intensity levels of the original image, g min and g max are the minimum and maximum intensity levels of the normalized image, and go (i, j) and g (i, j) are the gray levels at the coordinates (i, j) before and after normalization.

### 3.3 Median filter

Pre-processing is an important issue in low-level image processing. Using filtering it is possible to filter out the noise present in image. A high pass filter passes the frequent changes in the gray level and a low pass filter reduces the frequent changes in the gray level of an image. That is; the low pass filter smoothes and often removes the sharp edges. A special type of low pass filter is the Median filter. The Median filter takes an area of image (3 x 3, 5 x 5, 7 x 7, etc), observes all pixel values in that area and puts it into the array called element array. Then, the element array is sorted and the median value of the element array is found out. We have achieved this by sorting the element array in the ascending order using bubble sort and returning the middle elements of the sorted array as the median value. The output image array is the set of all the median values of the element arrays obtained for all the pixels [13]. Median filter goes into a series of loops which cover the entire image array.

Following are some of the important features of the Median filter: It is a non-linear digital filtering technique. It works on a monochrome color image. It reduces "speckle" and "salt and paper" noise. It is easy to change the size of the Median filter. It removes noise in image, but adds small changes in noise-free parts of image. It does not require convolution. Its edge preserving nature makes it useful in many cases.

The median value selected will be exactly equal to one of the existing brightness value so that no round-off error is involved when we work independently with integer brightness values comparing to the other filters [13, 14]. Fig 1. shows the original image.

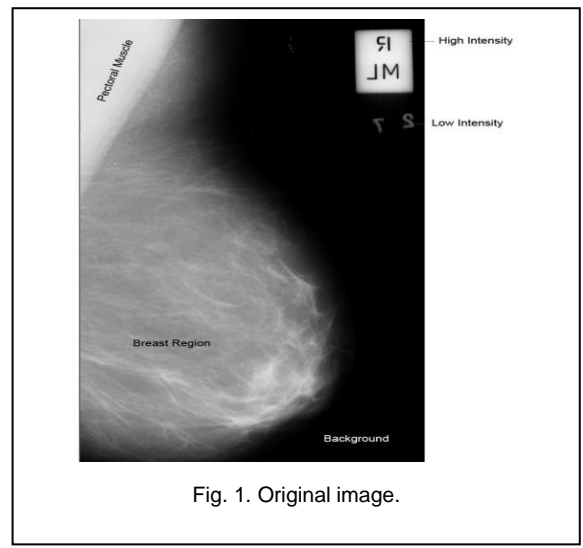

Fig. 1. Original image.

## 4 PROPOSED APPROACH

### 4.1 Connected component labeling

Connected component labeling is one of the most common operations in virtually all image processing applications. In machine vision most objects have surfaces. The points in a connected component form a candidate region to represent an object. The image object, which is the component, is separated from the background image on binary image. Then each component is labeled and displayed as output images. Points belonging to a surface project to spatially closed points. The notion of 'spatially closed' is captured by connected components in digital images [16]. Fig 2 shows the 8-neighborhood connected component.

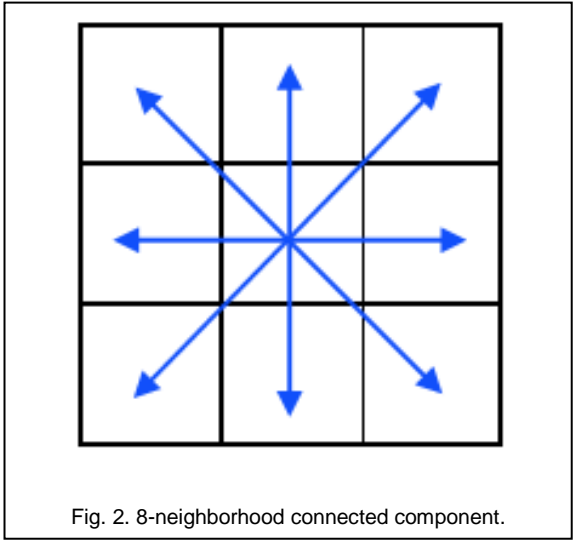

Fig. 2. 8-neighborhood connected component.

In binary valued digital imaging, a pixel can either have a value of 1 when it's part of the pattern or 0 when it's part of the background, there is no grayscale level. (We will assume that pixels with value 1 are black while zero valued pixels are white).

In order to identify objects in a digital pattern, we need to locate groups of black pixels that are "connected" to each other. In other words, the objects in a given digital pattern are the connected components of that pattern.

In general, a connected component is a set of black pixels, P, such that for every pair of pixels pi and pj in P, there exists a sequence of pixels pi, ..., pj such that:
a) All pixels in the sequence are in the set P i.e. are black, and
b) Every 2 pixels that are adjacent in the sequence are "neighbours".

**Proposed algorithm:**

HISNCCLM (Im, th)
Im- original image. Th- value

(1) $N\,Im \leftarrow Im$
(2) $chek(Im)\,\text{left or right}$
(3) If left
  $Im \leftarrow fliplr(Im)$
  End
(4) $hist \leftarrow hist(Im)$
(5) $f\,Im \leftarrow medifilt\,2(Im)$
(6) $bw\,Im \leftarrow im2bw(\,f\,Im,th\,)$
(7) $l \leftarrow bwlabel(\,bw\,Im,8\,)$
(8) $X \leftarrow find(\,l == 1\,)$ // breast region extraction
  $X \leftarrow find(\,l \sim= 1\,)$ // pectoral muscle removal
(9) $bw1 = f\,Im$
(10) $bw1(\,X\,) = 0$
  $return(\,bw1\,)$

## 5 WORKED EXAMPLE

In this section, the mammogram image, identification number mdb075 from the MIAS database is considered. This mammogram image is malignant, fatty image and it contains asymmetry masses. It is a left side breast image. It will be turn over in to right view. Fig 3(a) and 3(b). shows the proposed process.

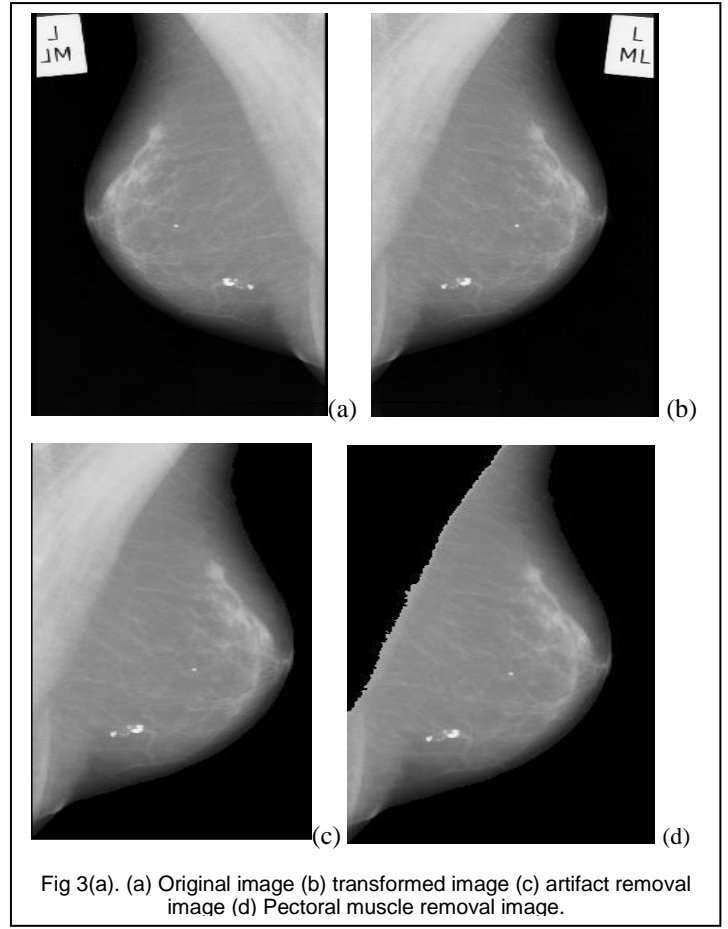

Fig 3(a). (a) Original image (b) transformed image (c) artifact removal image (d) Pectoral muscle removal image.

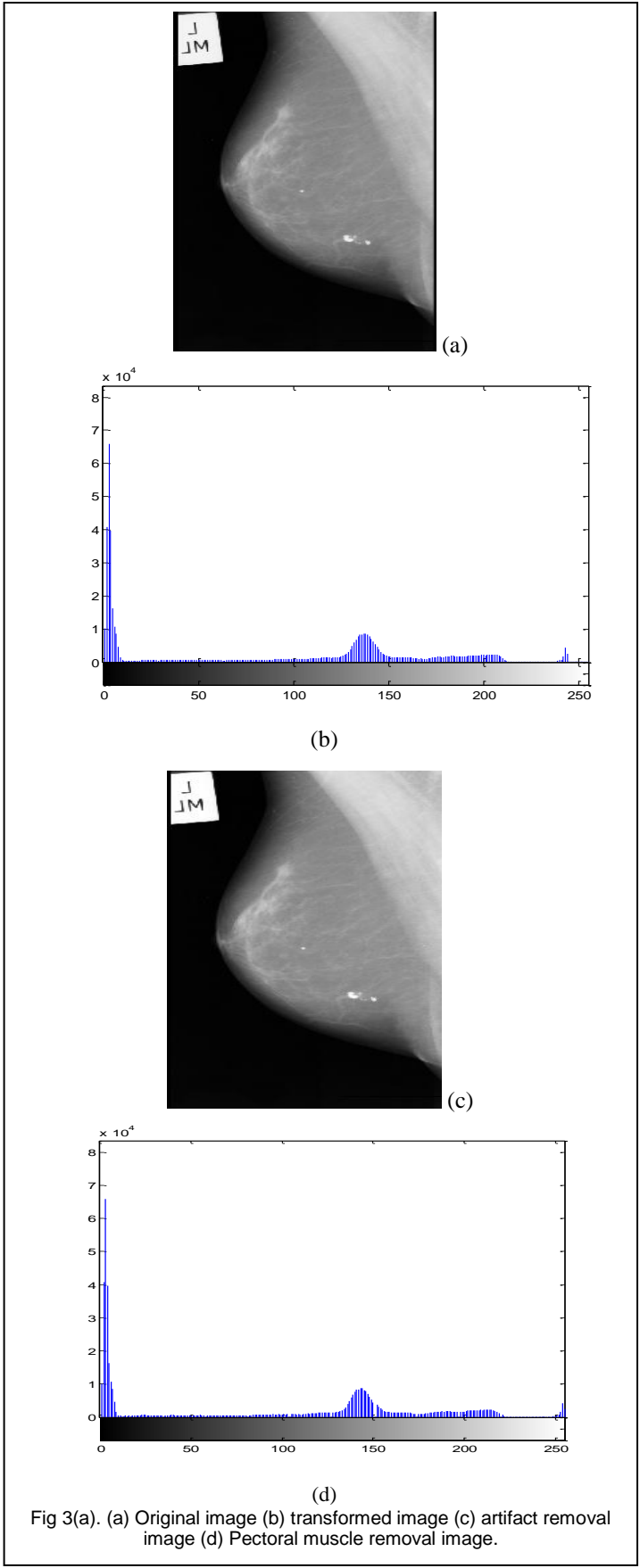

Fig 3(a). (a) Original image (b) transformed image (c) artifact removal image (d) Pectoral muscle removal image.

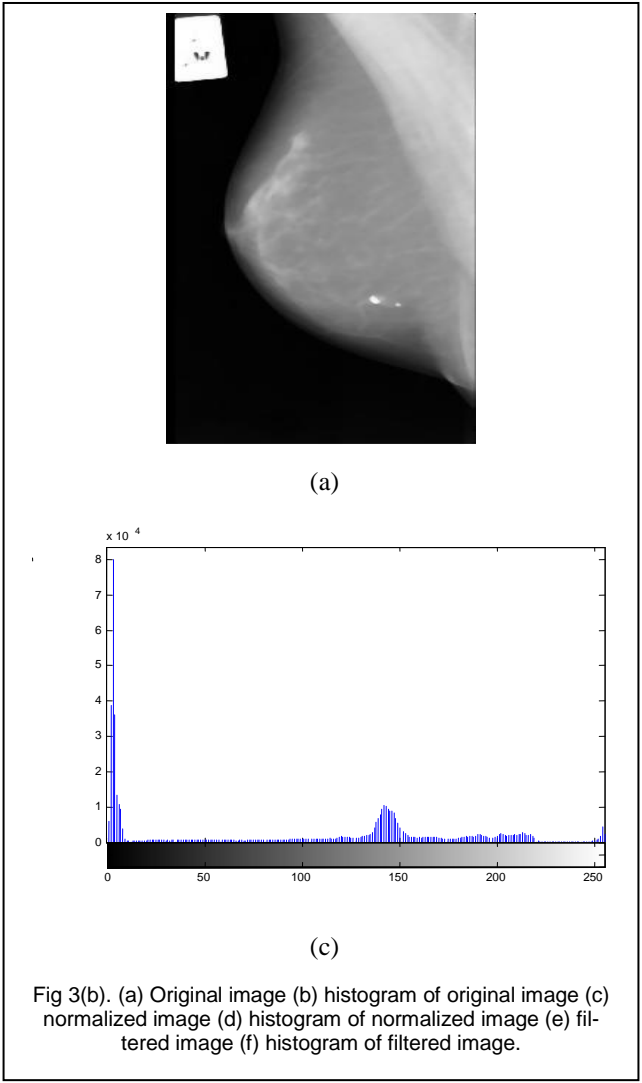

Fig 3(b). (a) Original image (b) histogram of original image (c) normalized image (d) histogram of normalized image (e) filtered image (f) histogram of filtered image.

## 6 EXPERIMENTAL RESULTS
### 6.1 Artifact removal

Raw mammogram image contains wedges and labels; these may produce unnecessary disturbances during mass detection process. Hence, it should be removed. In the proposed method, histogram based 8-neighborhood connected component labelling method algorithm are used to remove these artifacts.

At the initial phase of the pre-processing the image will be reviewed whether right view or left view, if it is in left view, it will be turn over in to right view. Then Median filtered gray scale image should be converted in to binary image based on thresholding. From Label connected components the pixels labeled 1 will be picked up as large component, the rest of the component are removed. From the above process smaller artifact regions can be easily removed from the mammogram. Fig 4. Shows Artifact removal image.

| Image Id | Original image With Artifact | Binary image | Binary Image of the Breast Region | Breast Region after removing artifact |
|---|---|---|---|---|
| Mdb096 | 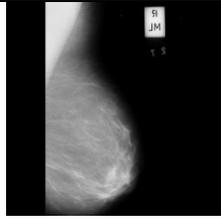 | 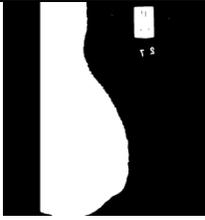 | 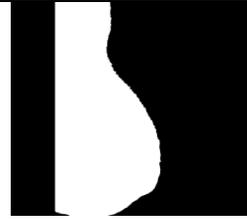 | 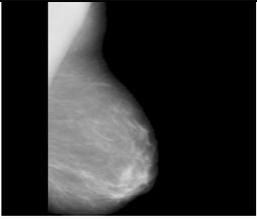 |
| Mdb201 | 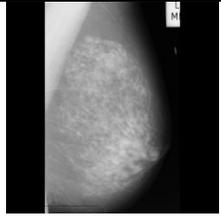 | 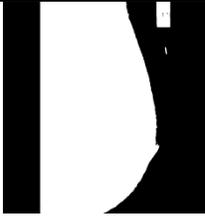 | 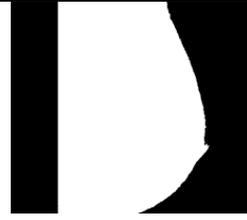 | 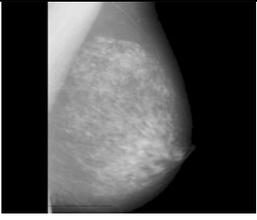 |
| Mdb012 | 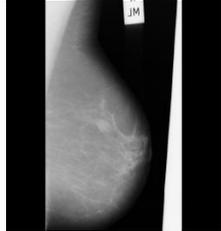 | 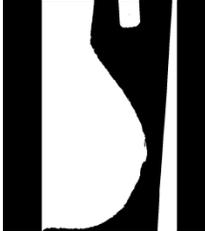 | 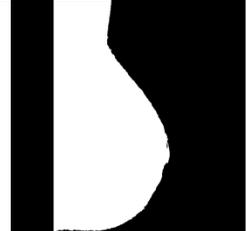 | 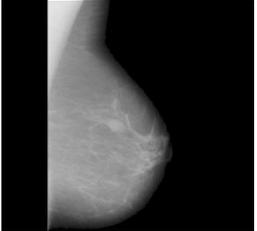 |
| Mdb058 | 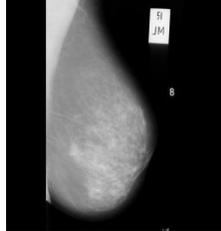 | 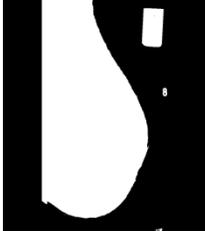 | 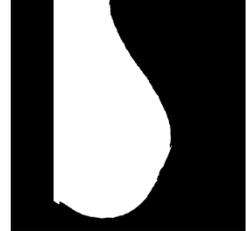 | 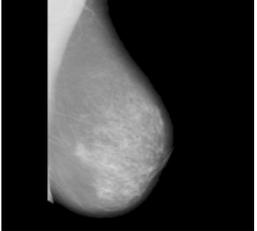 |
| Mdb132 | 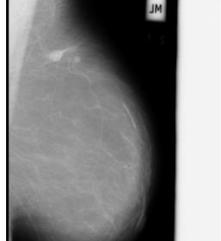 | 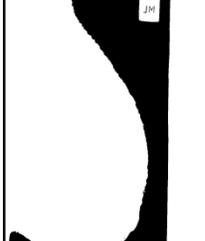 | 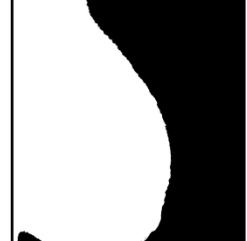 | 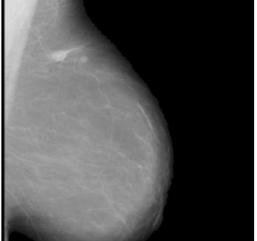 |
| Mdb272 | 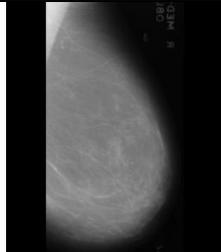 | 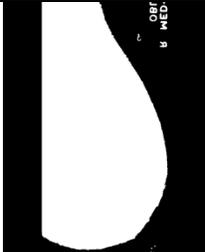 | 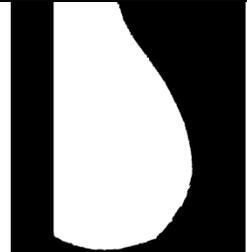 | 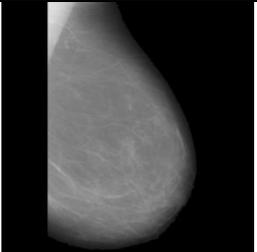 |

Fig 4. Results of Artifact.

| Image Id | Breast Region after removing artefact | Binary image of the pectoral region | Original pectoral region image | Breast Region after removal of Pectoral region |
|---|---|---|---|---|
| Mdb096 | 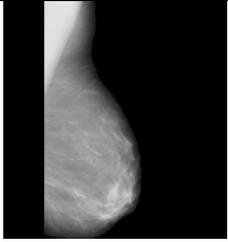 | 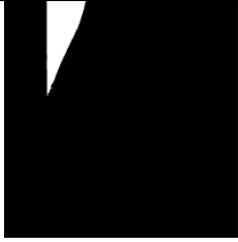 | 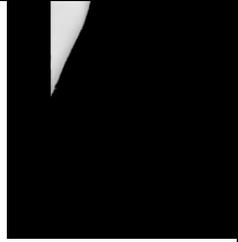 | 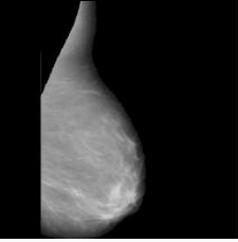 |
| Mdb201 | 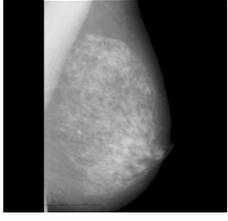 | 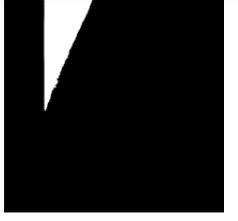 | 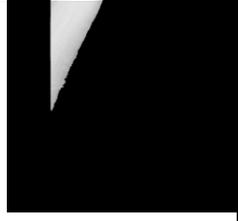 | 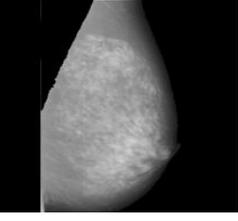 |
| Mdb012 | 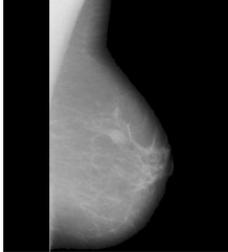 | 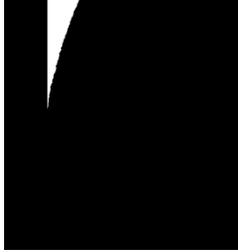 | 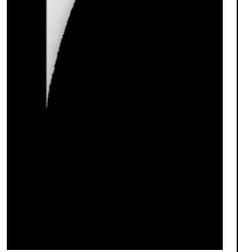 | 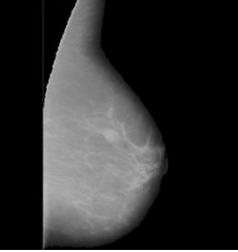 |
| Mdb058 | 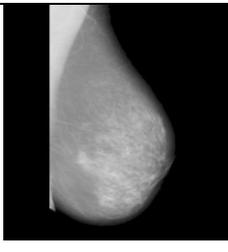 | 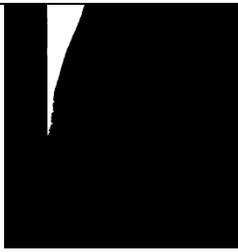 | 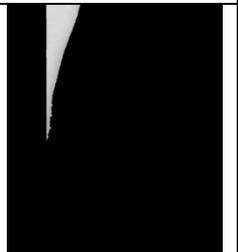 | 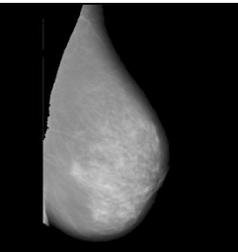 |
| Mdb132 | 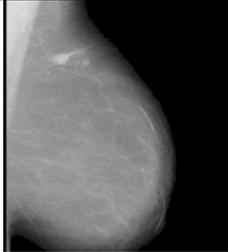 | 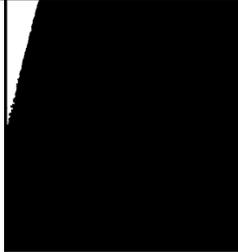 | 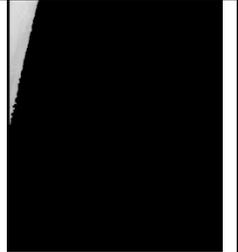 | 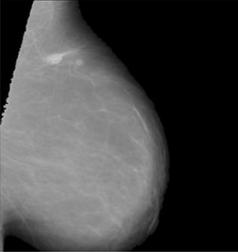 |
| Mdb272 | 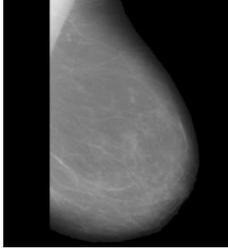 | 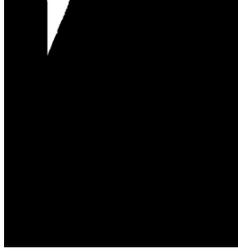 | 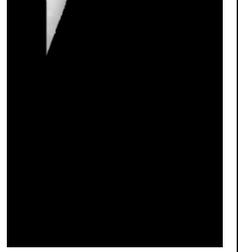 | 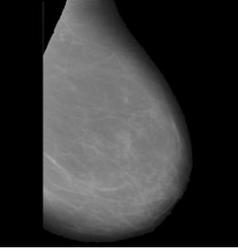 |

Fig 5. Results of Pectoral Region Removal.

Table 1. Comparative analysis of pectoral muscle removal

| Author reference | Methods | Images used for experiment | Acceptable (%) | Unacceptable (%) |
|---|---|---|---|---|
| **Mustra et al. [17] 2009** | Hybrid with Bit depth reduction and wavelet decomposition | 40 | 34(85%) | 6(15%) |
| **Li Liu et al.[20] 2011** | AD method | 100 | 81(81%) | 19(19%) |
| **Raba et al. [18] 2005** | Adaptive histogram approach | 322 | 278(86%) | 44(14%) |
| **Chen et al. [19] 2010** | Histogram thresholding, edge detection in scale space, contour growing and polynomial fitting | 86 | 80(93.5%) | 6(6.5%) |
| **Proposed** | Histogram based 8-neighborhood connected component labelling method | **322** | **288 (89.5)** | **34 (10.5)** |

## 6.3 Breast Region Extraction

The pectoral is the term relating to the chest. It is a large fan shaped muscle that covers much of the front upper chest. Hence during the mammogram capturing process pectoral muscle also would be captured. The pectoral muscle represents a predominant density region. Hence it will severely affect the result of image processing. For better detection accuracy pectoral region should be removed from mammogram image. The orientation of the breast should be found out to remove the pectoral region. After the removing the artifact, the pectoral region also removed using connected component labeling methods. Fig 5. Shows the pectoral muscle removal image. Table 1 show the comparative analysis of pectoral muscle removal results. For the 322 mammograms evaluated, the mean values of accuracy and error are 0.894 and 0.106 respectively.

## 7 CONCLUSION

The proposed histogram based 8-neighborhood connected component labelling method for breast region extraction and removal of pectoral muscle. The results obtained over MIAS database show excellent output. This algorithm is used to detect pectoral muscle accurately and suppress the pectoral muscle successfully without losing any information from the rest of the mammogram. Further, the resultant mammogram can be used further for the detection abnormalities in human breast like calcification, circumscribed masses, spiculated masses and other ill-defined masses, circumscribed lesions, asymmetry analysis etc. This algorithm has the potential for further development because of its simplicity and it also encourages results that will motivate real-time breast cancer diagnosis system.

## Acknowledgment

The author immensely acknowledges the UGC, New Delhi for partial financial assistance under UGC-SAP (DRS) Grant No. F.3-50/2011.

The first and third authors immensely acknowledge the partial financial assistance under University Research Fellowship, Periyar University, Salem